\documentclass{article}

\usepackage[nonatbib,preprint]{nips_2018}

\usepackage[utf8]{inputenc}
\usepackage[T1]{fontenc}
\usepackage{hyperref}
\usepackage{url}
\usepackage{booktabs}
\usepackage{amsfonts}
\usepackage{nicefrac}
\usepackage{microtype}
\usepackage{natbib}
\usepackage{amsmath}
\usepackage{amssymb}
\usepackage{makecell}
\usepackage{graphicx}
\usepackage{xspace}
\usepackage{paralist}
\usepackage{comment}
\usepackage{mathtools}
\usepackage{algorithm}
\usepackage{algorithmic}
\usepackage{xcolor}
\usepackage{ragged2e}
\usepackage{subcaption}
\usepackage{commath}
\usepackage[capitalise,noabbrev,nameinlink]{cleveref}

\usepackage{xparse}
\usepackage{mathtools}
\usepackage{etoolbox}
\usepackage{setspace}
\usepackage[symbol]{footmisc}

\setcitestyle{authoryear,round,citesep={;},aysep={,},yysep={;}}

\graphicspath{{imgs/}}

\renewcommand{\norm}[1]{\left\lVert#1\right\rVert}

\DeclarePairedDelimiterX{\SquareBrackets}[1]{[}{]}{#1}
\DeclarePairedDelimiterX{\RoundBrackets}[1]{(}{)}{#1}
\DeclarePairedDelimiterX{\DivergenceBrackets}[2]{[}{]}{#1\;\delimsize\|\;#2}

\NewDocumentCommand{\pr}{ O{p} r() }{
  \def\prArg{#2}\patchcmd{\prArg}{|}{\mid}{}{}#1\RoundBrackets{\prArg}}
\NewDocumentCommand{\p}{ r() }{\pr[p](#1)}
\NewDocumentCommand{\q}{ r() }{\pr[q](#1)}
\NewDocumentCommand{\Normal}{ r() }{\pr[\operatorname{Normal}](#1)}
\NewDocumentCommand{\Cat}{ r() }{\pr[\operatorname{Cat}](#1)}
\NewDocumentCommand{\Bin}{ r() }{\pr[\operatorname{Bin}](#1)}
\NewDocumentCommand{\Beta}{ r() }{\pr[\operatorname{Beta}](#1)}
\NewDocumentCommand{\Bernoulli}{ r() }{\pr[\operatorname{Bernoulli}](#1)}
\NewDocumentCommand{\Dir}{ r() }{\pr[\operatorname{Dir}](#1)}

\usepackage{hyperref}
\usepackage{url}

\title{Modulated Policy Hierarchies}

\author{
  Alexander Pashevich\\
  Inria\thanks{Univ. Grenoble Alpes, Inria, CNRS, INPG, LJK, Grenoble, France}\\
  Grenoble, France\\
  \makebox[5em][c]{\texttt{alexander.pashevich@inria.fr}} \\
  \And
  Danijar Hafner\\
  Google Brain\\
  Toronto, ON, Canada\\
  \texttt{mail@danijar.com}\\
  \And
  James Davidson\\
  Third Wave Automation\\
  Union City, CA, USA \\
  \texttt{james@thirdwave.ai}\\
  \And
  Rahul Sukthankar\\
  Google Research\\
  Mountain View, CA, USA\\
  \texttt{sukthankar@google.com}\\
  \And
  Cordelia Schmid\\
  Inria\footnote[1]{}\\
  Grenoble, France\\
  \texttt{cordelia.schmid@inria.fr}\\
}

\begin{document}

\maketitle

\begin{abstract}
Solving tasks with sparse rewards is a main challenge in reinforcement learning. While hierarchical controllers are an intuitive approach to this problem, current methods often require manual reward shaping, alternating training phases, or manually defined sub tasks. We introduce modulated policy hierarchies (MPH), that can learn end-to-end to solve tasks from sparse rewards. To achieve this, we study different modulation signals and exploration for hierarchical controllers. Specifically, we find that communicating via bit-vectors is more efficient than selecting one out of multiple skills, as it enables mixing between them. To facilitate exploration, MPH uses its different time scales for temporally extended intrinsic motivation at each level of the hierarchy. We evaluate MPH on the robotics tasks of pushing and sparse block stacking, where it outperforms recent baselines.
\end{abstract}

\section{Introduction}

\suppressfloats  
\begin{figure}[t]
\minipage{0.24\textwidth}
  \includegraphics[height=3.8cm,width=\linewidth]{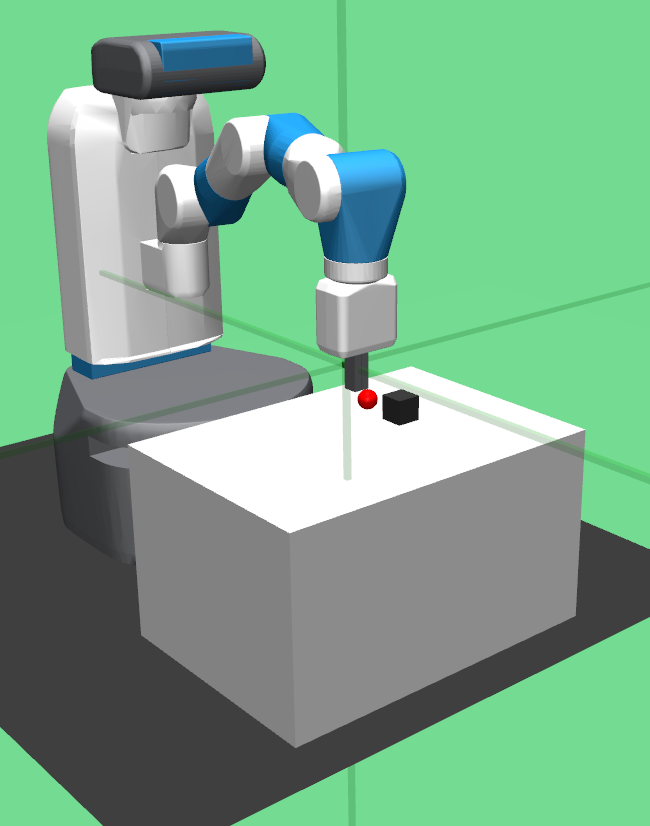}
\endminipage\hfill
\minipage{0.24\textwidth}
  \includegraphics[height=3.8cm,width=\linewidth]{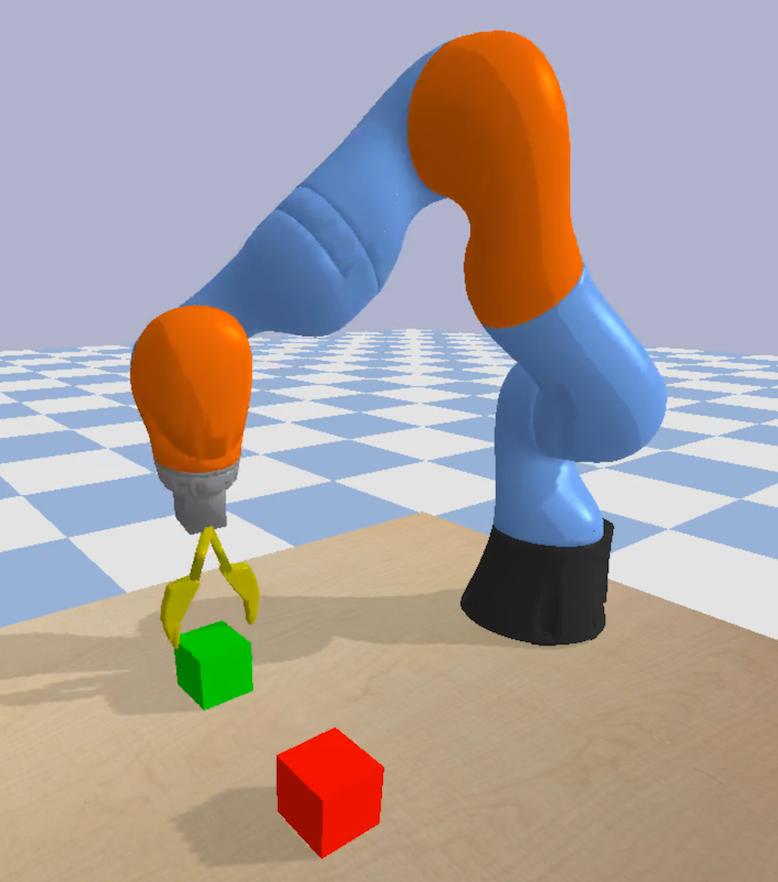}
\endminipage\hfill
\minipage{0.24\textwidth}
  \includegraphics[height=3.8cm,width=\linewidth]{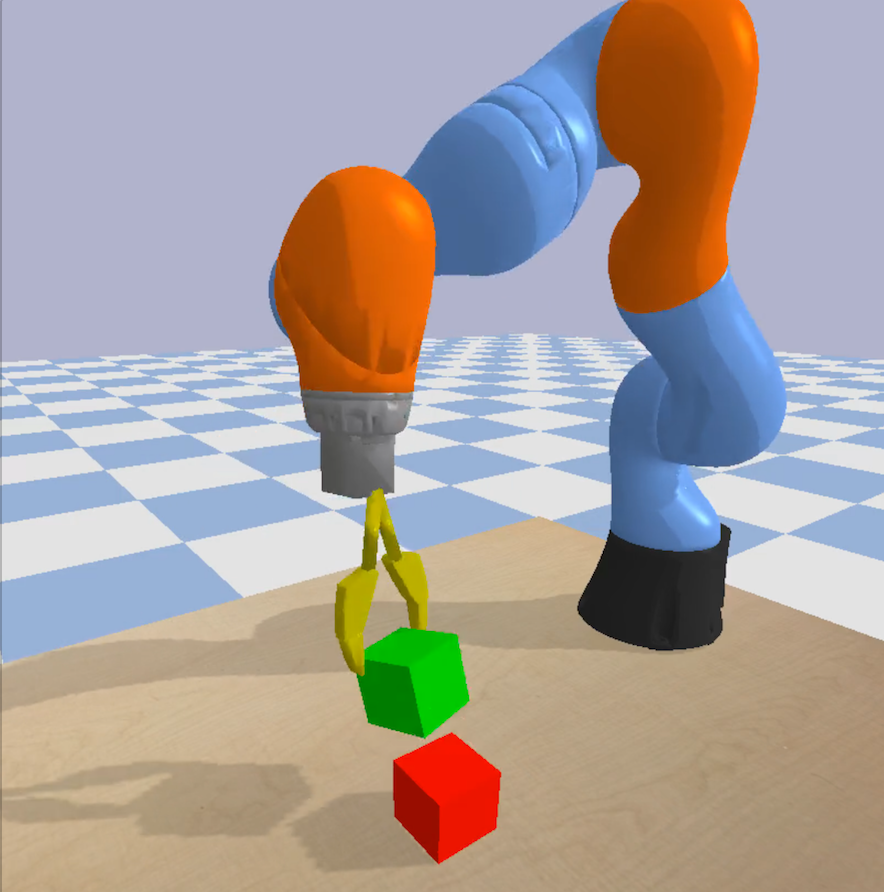}
\endminipage\hfill
\minipage{0.24\textwidth}%
  \includegraphics[height=3.8cm,width=\linewidth]{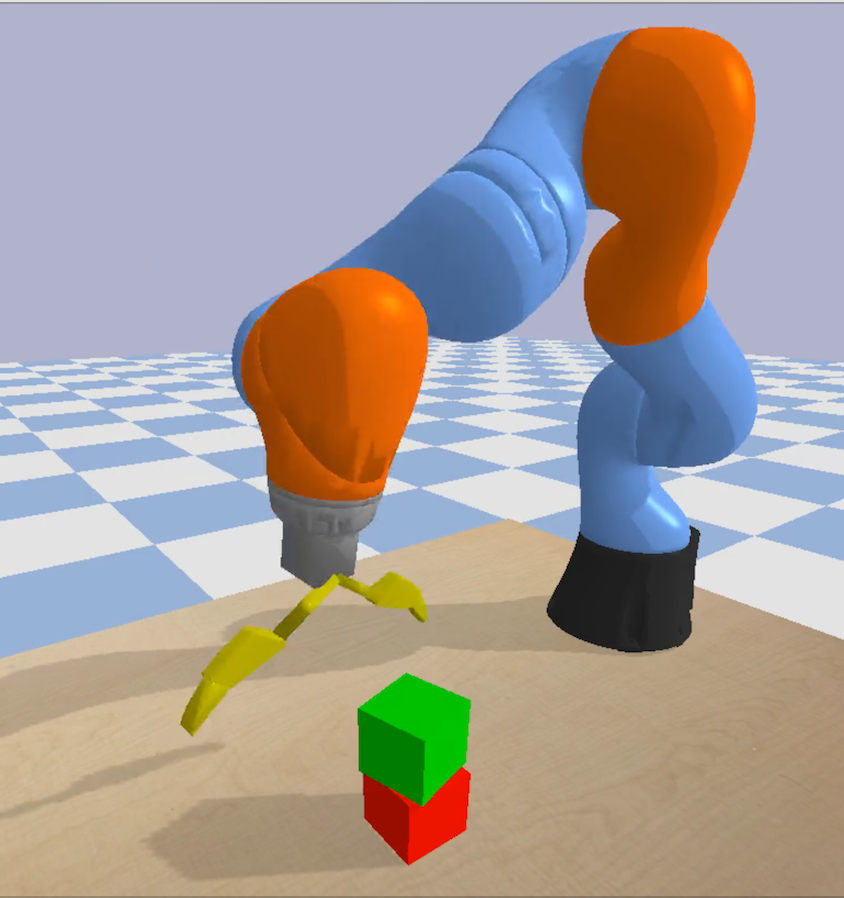}
\endminipage
\caption{Manipulation tasks with only sparse rewards using our hierarchical agent.
Left: FetchPush-v1 environment.
Right: Block stacking environment. The agent learns to reach the first block, grasp it, and stack it on top of the other block. The corresponding modulation signals for this policy are visualized in \cref{fig:skills}.}
\label{fig:envs}
\end{figure}

Reinforcement learning (RL) has shown impressive results on artificial tasks such as game playing \citep{Mnih2015Human-levelLearning,Silver2016MasteringSearch}, where collecting experience is cheap. However, for robotics tasks, such as locomotion and manipulation \citep{Kober2013,Gu2016}, current algorithms often require manually designed smooth reward functions, limiting their applicability in real-world scenarios. In this paper, we approach learning from sparse rewards using hierarchical reinforcement learning (HRL), where multiple levels of temporally-abstract controllers modulate another to produce an action. We propose a novel hierarchical agent that is simple to train and learns to push objects and stack blocks end-to-end from sparse rewards, as shown in \cref{fig:envs}. To achieve this, we consider three common challenges of HRL.

\paragraph{Stability.}

Simultaneously updating the levels of a hierarchical agent introduces non-stationarity since the levels affect another, resulting in unstable learning dynamics. Prior HRL algorithms thus often introduce multiple training phases to stabilize learning \citep{Heess2016LearningControllers,MetaLearningSharedHierarchies}. This requires more effort in implementation and introduces additional hyperparameters, which may in part prevent HRL from becoming standard for RL problems. We alleviate this problem by jointly training the levels as separate PPO agents \citep{PPO}, encouraging smooth changes in all levels of the hierarchy. We hope that the simplicity of this solution helps to make HRL more generally applicable.

\paragraph{Modulation.}

A critical design choice of hierarchical agents is the form of communication between levels. Typically, each level receives a modulation signal from the more abstract level above it \citep{Dayan1993FeudalLearning}. Such signal could be a categorical variable called an option \citep{sutton1999between,MetaLearningSharedHierarchies,Florensa2017StochasticLearning} or a continuous-valued activation vector \citep{Heess2016LearningControllers,Vezhnevets2017FeUdalLearning,LatentSpacePolicies}. While a categorical signal allows to select exactly one skill at a time, a continuous signal allows smooth modulation over lower levels. Inspired by this trade-off, we propose communication via bit-vectors, which allows to mix multiple skills. Empirically, this outperforms categorical modulation signals.

\paragraph{Exploration.}

While hierarchical controllers with different time-scales have a built-in prior for temporally extended behavior, this does not necessarily help the exploration of skills \citep{DIAYN}. For this reason, HRL methods often report transfer performance after pre-training the agent or part of it on manually defined subtasks \citep{Heess2016LearningControllers,tessler2017dsn,kulkarni2016hierarchical}. Intrinsic motivation or curiosity \citep{schmidhuber2010formal,pathak2017curiosity} is a common approach to exploration, but is not commonly used for hierarchical agents. We achieve temporally extended exploration by employing intrinsic motivation at each level of the hierarchy, resulting in an agent that learns from sparse rewards without pre-training on simpler tasks.

We summarize the main contributions of this paper as follows:
\begin{itemize}
\item We introduce modulated policy hierarchies (MPH), a hierarchical agent that is trained jointly without requiring pre-training tasks or multiple training phases.
\item We model modulation signals as bit vectors instead of the typical one-hot modulation, allowing the agent to interpolate between its skills.
\item We employ intrinsic motivation based on prediction error of dynamics models on all levels of the hierarchy, resulting in temporally extended exploration.
\item We evaluate our method together with recent HRL algorithms on pushing and sparse block stacking and provide ablation studies of the design choices.
\end{itemize}

\section{Related work}
\label{sec:related_work}

\paragraph{Manipulation.}

Learning algorithms have been applied to a variety of robotics tasks, such as grasping~\citep{Pinto2016,Lampe2013AcquiringLearning}, opening bottles using supervised learning~\citep{Levine2015End-to-EndPolicies}, learning to grasp with reinforcement learning~\citep{Levine2016LearningCollection}, opening doors~\citep{Gu2016}, and stacking Lego blocks~\citep{Popov2017Data-efficientManipulation}. Here, we examine a pushing task, FetchPush-v1, and a stacking task similar to the one described by \citet{Popov2017Data-efficientManipulation}. We focus on pushing and stacking tasks as they conceptually require subskills, such as reaching and grasping. Previously, solving block stacking required to manually design a smooth reward function~\citep{Popov2017Data-efficientManipulation} or imitation learning~\citep{2017arXiv170307326D} with an existing low-level controller that could already grasp and place blocks.

\paragraph{Stability.}

HRL inherits common instability issues of RL~\citep{DeepRLThatMatters}. Moreover, training multiple controllers simultaneously can lead to degenerate solutions~\citep{bacon2017option}. Pre-training the low levels ~\citep{Heess2016LearningControllers} or alternating training between levels~\citep{MetaLearningSharedHierarchies} has been proposed to improve learning stability. Other methods add regularization terms based on entropy~\citep{bacon2017option,hausman2018learning,LatentSpacePolicies} or mutual information~\citep{daniel2016hierarchical,Florensa2017StochasticLearning}. We find that the slow change of the action distribution that PPO encourages can be a simple and practical way to mitigate instability. In addition to \citet{MetaLearningSharedHierarchies}, we find that we can train all levels simultaneously without degeneracies.

\paragraph{Modulation.}

A core design choice of HRL agents is how higher levels communicate with the levels underneath them. The options framework \citep{sutton1999between} uses a categorical signal that switches between low-level policies that can be implemented as separate networks \citep{tessler2017dsn,MetaLearningSharedHierarchies}. This approach requires a large number of parameters and does not allow skills to share information. Another line of work is based on feudal learning~\citep{Dayan1993FeudalLearning}, where a typically continuous valued signal modulates the lower level. In this context, the modulation signal is often referred to as goal. It can be specified directly in the observation space~\citep{hiro2018arxiv} or in a learned embedding space~\citep{Vezhnevets2017FeUdalLearning,kulkarni2016hierarchical,Heess2016LearningControllers}. However, such methods usually require prior knowledge in a form of pre-training~\citep{hausman2018learning} or reward shaping~\citep{LatentSpacePolicies} which our method with binary signals does not need.

\paragraph{Exploration.}

One of the major goals of HRL is to address hard exploration problems with long-horizon and sparse rewards. Structured exploration provides a mechanism to more effectively guide exploration and is usually referred to as intrinsic motivation. The intrinsic motivation methods vary from curiosity-based bonuses~\citep{houthooft2016vime,pathak2017curiosity} to state visitation counters~\citep{BellemareSOSSM16, OstrovskiBOM17}. However, such approaches have only been explored for single-level policies while one can imagine taking advantage of the intrinsic motivation at both layers of the hierarchy.

\section{Modulated Policy Hierarchy}
\label{sec:approach}

\begin{figure}
\centering
\hspace*{3em}
\begin{subfigure}[t]{.22\textwidth}
\includegraphics[width=\textwidth]{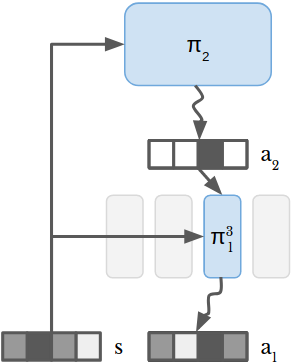}
  \caption{Options}
  \label{fig:models_options}
\end{subfigure}\hfill
\begin{subfigure}[t]{.22\textwidth}
\includegraphics[width=\textwidth]{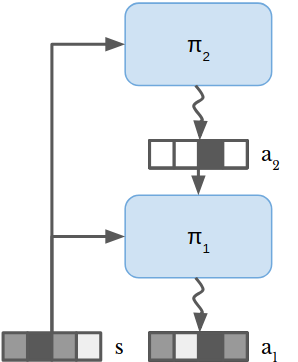}
  \caption{One-hot modulation}
  \label{fig:models_onehot}
\end{subfigure}\hfill
\begin{subfigure}[t]{.22\textwidth}
\centering
\includegraphics[width=\textwidth]{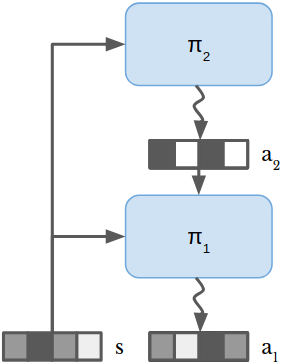}
  \caption{MPH (ours)}
  \label{fig:models_mph}
\end{subfigure}\hspace*{3em}
\caption{Overview of hierarchical policies.
(a)~The options agent selects between separate skill networks using a categorical master policy.
(b)~The one-hot agent combines the skills into a single network and is modulated by a 1-hot signal.
(c)~Our modulated policy hierarchy sends a binary vector, allowing for richer communication and mixing of skills.}
\label{fig:models}
\end{figure}

\paragraph{Preliminaries.}

We follow the typical formulation of reinforcement learning as Markov decision process. At each discrete time step $t$, the agent receives state $s_t$ from the environment, selects an action $a_t$, receives a scalar reward $r_t$, and transitions into a new state $s_{t+1}$. We aim to maximize the expected return $\mathbb{E}_{\pi} \sum_{k=0}^{\infty} \gamma^k r_{t+k}$ using a policy $\pi: \mathcal{S}\times\mathcal{A}\to [0,1]$. The policy is parameterized with parameters $\theta$ and denoted as $\pi_{\theta}(a_t|s_t)$, thus at each timestep the chosen action is given by $a_t \sim \pi_{\theta}(\cdot | s_t)$.

\paragraph{Hierarchical controller.}

MPH learns a hierarchy of policies $\Pi = \{\pi_1, \pi_2, \ldots, \pi_n\}$, where in our experiments we consider two-level policies ($n=2$). MPH policies have their own state and action spaces $\pi_k: \mathcal{S}^k\times\mathcal{A}^k\to [0,1]$. Each policy is represented with a single network and modulated by bit vectors from the policies above (\cref{fig:models_mph}). In contrast, the options framework switches between independent skill networks by a categorical modulation signal (\cref{fig:models_options}). The categorical signal might be also used with the skill policies merged into a single network (\cref{fig:models_onehot}).

\paragraph{Modulation signal.}

The highest-level (master) policy $\pi_n$ receives the state from the environment ($\mathcal{S}^n = \mathcal{S}$) and outputs a bit vector of size $m_n$ as its action. Each intermediate-level policy $\pi_k$ receives the environment state concatenated with the modulation signals from the layers above. The policies are implemented as fully-connected neural networks that predict the probabilities of the bits. Given the probabilities, the modulation signal is generated by sampling from $m_k$ independent Bernoulli distributions. Once sampled, the signal is passed to all lower policies. Finally, the lowest level policy $\pi_1$ (worker) observes all the modulation signals and the state. In the two-level structure, $\pi_1$ receives the environment state and the master modulation, $s^1 \in \mathcal{S}^1 = \mathcal{S} \cup \mathcal{A}^2$. The worker policy outputs the final action $a^1 \in \mathcal{A}^1 = \mathcal{A}$ which is applied to the environment.

\paragraph{Time scales.}

To encourage each level in the hierarchy to attend to different time-scales, we activate higher-level policies less frequently, i.e., $T_{k+1} > T_k$ where $T_k$ is the time-scale of the policy at level $k$. When a policy is inactive in a given time step, it outputs the same modulation signal as was generated in the previous time step, which promotes consistency in higher-level decisions and facilitates longer-term planning. The policies at each level only receive inputs at time steps for which they are active.

\paragraph{Optimization.}

We train MPH policies using PPO \citep{PPO} which is a state-of-the-art on-policy method. PPO guarantees that after each update the new policy does not deviate too far from the old one in terms of KL divergence. We use PPO for each layer independently which also means that the MDPs seen by high-level policies change during training due to low-level policies updates. However, given the PPO guarantees, we can ensure that after a training step, the MDP on each layer of the hierarchy remain close to the old MDP in terms of transition probabilities change. As a result, the optimization problem solved by PPO for higher layers changes smoothly during the updates.  This fact makes MPH more stable to train than most HRL approaches. Please refer to \cref{appendix:derivations} for exact bounds and full derivation.

\paragraph{Hierarchical exploration.}
\label{sec:curiosity}

Since MPH is designed for environments with sparse reward signals, we employ intrinsic motivation to accelerate learning. As suggested by \cite{pathak2017curiosity}, we add intrinsic motivation to our agent in the form of a curiosity-driven exploration bonus. We apply this independently for each level of the hierarchy and on the corresponding time-scale. In practice, this means that higher-level policies, which operate on longer time-scales, are more curious about longer term effects than lower-level policies. The reward for a policy at level $k$ is defined as
\begin{equation}
    R^k_t = R^\text{env}_t + \norm{\hat{\phi}^F_k({s}^k_{t+1}) - \phi_k(s^k_{t+1})}_2,
\end{equation}
where $\phi_k$ is a learned embedding and $\hat{\phi}^F_k(s^k_{t+1})$ is a prediction of the next state $s^k_{t+1}$ given $(s^k_t, a^k_t)$. The standard method for learning $\phi_k$ requires an inverse model for the action prediction, but we find that training a \emph{reverse} model instead works better. Specifically, the reverse model predicts the previous state $s^k_{t}$ given $(s^k_{t+1}, a^k_{t})$. To learn the embedding, we jointly train forward and reverse models by minimizing the loss
\begin{equation}
    L = \beta \norm{\hat{\phi}^F_k({s}^k_{t+1}) - \phi_k(s^k_{t+1})}_2 + (1-\beta) \norm{\hat{\phi}^R_k(s^k_{t}) - \phi_k(s^k_{t})}_2 - \lambda (\norm{\phi_k(s^k_{t})}_1 + \norm{\phi_k(s^k_{t+1})}_1),
\end{equation}%
where we add a regularization term to prevent trivial embeddings $\phi_k$ to be learned; $\beta \in (0,1)$ is a scalar weighting the reverse model loss against the forward model loss, and $\lambda \in \mathbb{R}$ is a regularization scaling factor.

\section{Experiments}
\label{sec:experiments}

We compare our approach to baselines and state-of-the-art approaches described in \cref{sec:baselines}. We evaluate our approach on two tasks with sparse rewards: block pushing and block stacking (see \cref{fig:envs}). First, we show that MPH outperforms the baselines on the block stacking in \cref{sec:stacking} and analyze the modulation signals produced by the master policy. Second, we compare MPH to baselines on the pushing task in \cref{sec:pushing}. Third, we show the benefits of temporally extended intrinsic motivation in \cref{sec:motivation}.

\subsection{Baselines}
\label{sec:baselines}

We compare to the following baselines:
\begin{itemize}
\item\textbf{PPO}\quad A flat policy trained using PPO.
\item\textbf{options}\quad An options hierarchy with separate skill networks corresponding to \cref{fig:models_options}.
\item\textbf{one-hot}\quad A two-level hierarchy with 1-hot modulation signal as in \cref{fig:models_onehot}.
\item\textbf{MLSH}\quad Meta Learning Shared Hierarchy by \citet{MetaLearningSharedHierarchies}.
\end{itemize}

All the hierarchies employ PPO as the core optimization method and use temporal abstraction for the master policy. We share the common hyperparameters between MPH and baselines for each task. We discuss the hyperparameters in more details in \cref{sec:stacking} and \cref{sec:pushing}. The last approach, MLSH~\citep{MetaLearningSharedHierarchies} is a recent, state-of-the-art approach that learns a set of skill policies, switched by a master policy. MLSH is trained stage-wise: a warm-up period where only the master is updated is alternated with a period where skills and master are trained jointly. We implemented the first three approaches and rely on the code of MLSH released by its authors.

\subsection{Stacking}
\label{sec:stacking}

\begin{figure}[t]
\centering
\begin{subfigure}[t]{.5\textwidth}
\includegraphics[width=\textwidth]{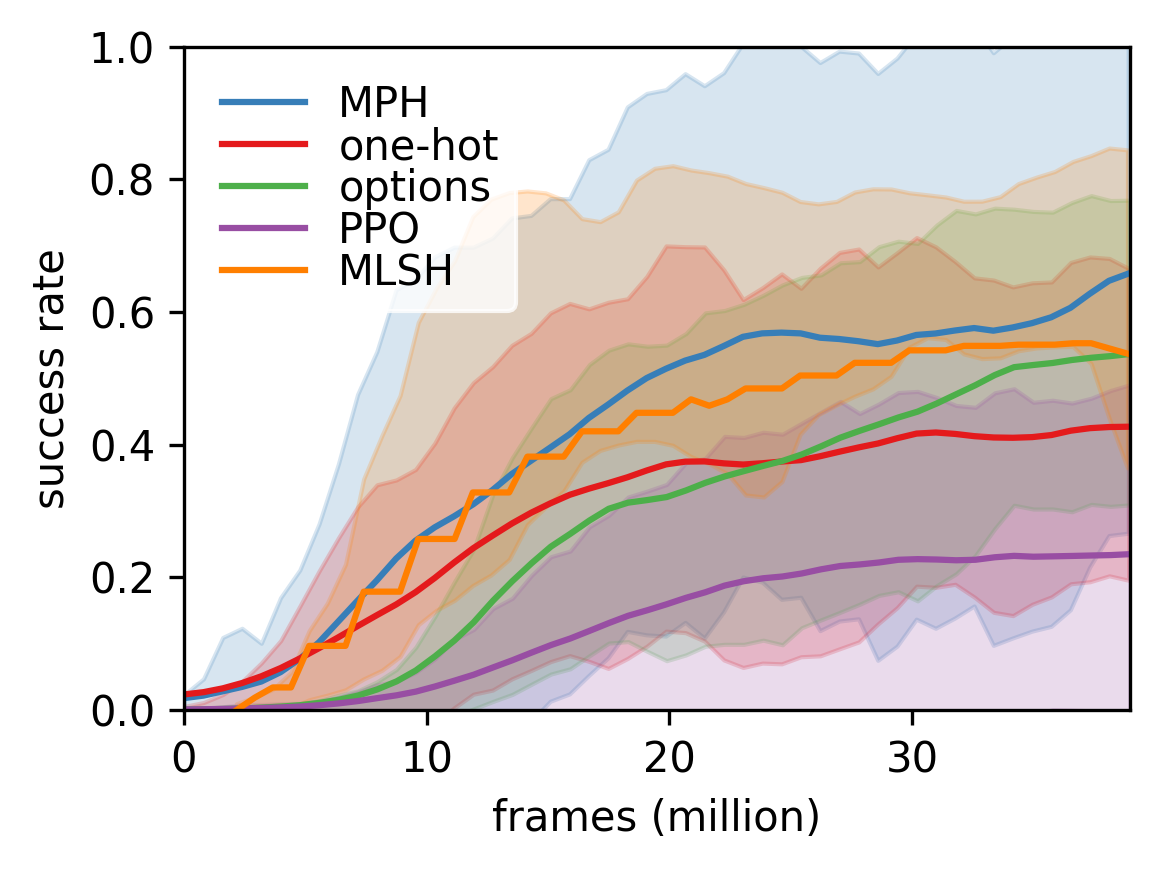}
\caption{Stacking}
\label{fig:stacking}
\end{subfigure}\hfill
\begin{subfigure}[t]{.5\textwidth}
\includegraphics[width=\textwidth]{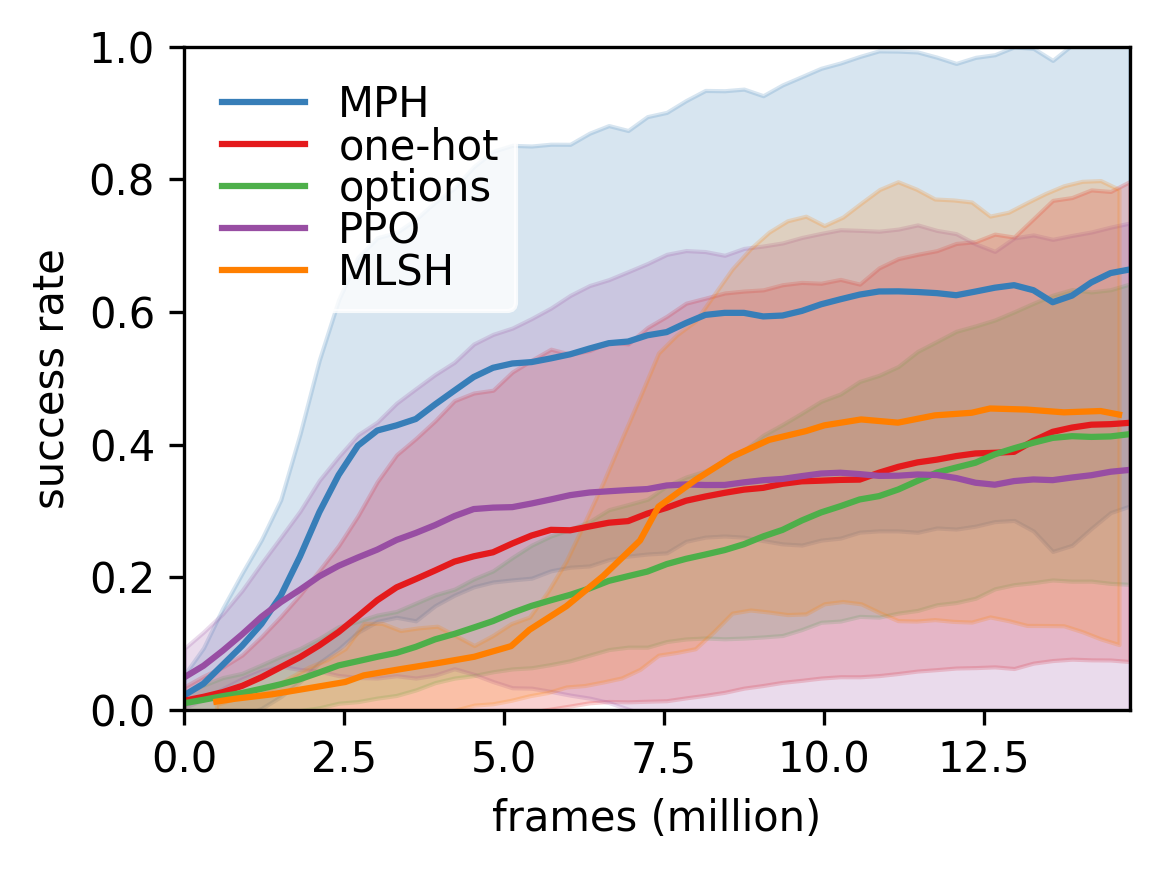}
\caption{FetchPush-v1}
\label{fig:pushing}
\end{subfigure}\hfill
\caption{Evaluation curves for MPH and baselines. The solid lines correspond to mean success rate, and the shaded lines show the standard deviation. For stacking, both values are mean values for 50 episodes and averaged over top 5 out of 16 random seeds. For pushing, we average over 32 episodes and use 5 random seeds.}
\end{figure}

\paragraph{Task description.}

The block stacking task is a \texttt{pybullet}~\citep{CoumansBulletEngine.} based simulation environment (see \cref{fig:envs} right). We use a model of the 7-DOF Kuka LBR arm with a 1-DOF pinch gripper. The scene contains two blocks and the goal is to stack one on top of the other. All episodes start in a randomly initialized state in terms of robot configuration and object placement. The state perceived by the agent consists of the angles and the angular velocities of the arm and the gripper, the tactile sensor for each finger of the gripper, location and orientation of each object in the scene as well as the relative distances of the two blocks to the pinch position of the gripper. The agent acts at the frequency of 40 Hz and outputs desired joint position change which is then applied using position control. The time horizon is 200 timesteps. The agent receives the following sparse rewards: a) for touching a block with the gripper, b) for successfully lifting a block, c) for holding a block above another one, and d) for placing a block on top of another block. We also reward the agent with a larger reward when the objects are stacked and the gripper pinch position is far enough from the tower.

\paragraph{Hyperparameters.}

We use identical network architectures and common hyperparameters for all the approaches including MPH. For the policies, value functions, and the models, we use fully connected neural networks with 2 hidden layers, consisting of 64 hidden units with tanh activation each. We use the implementation of PPO from \citet{hafner2017tfagents} and collect a batch of 50 rollouts using parallel environments. The learning rate is set to 0.0001, 0.01, and 0.005 for policies, value functions and models networks correspondingly. We use Adam as an optimizer. The maximum KL divergence step size of PPO is set to 0.001 and 0.002 for the master and the worker policies correspondingly. We update both the policies and the value networks using 40 training epochs. We set discount factor, $\gamma$, to 0.985 and the models loss coefficient, $\beta$, to 0.2. We use 3 as a width of all modulation signals and the number of skills for the baselines. For the master policy time-scale, we choose the best value among 4, 8, and 16. The options framework uses a time-scale of 8 for the master, the 1-hot baseline and MPH use the time-scale of 4. In the case of MLSH we adapt some of the parameters according to the suggestion of the authors: we use learning rates of 0.01 and 0.0003 for the master and the skill policies respectively, and use 10 groups of 12 cores to train MLSH, a warmup time of 20 (the best among 10, 20, 30), a training time of 70 (the best among 30, 50, 70), and a master policy time-scale of 25 (the best among 10, 25, 50).

\paragraph{Performance.}

\Cref{fig:stacking} shows that MPH outperforms the baselines on the stacking problem. We compare the success rates of the approaches averaged over 50 episodes. The stacking is considered successful if in the end of the episode the blocks are in a stacked configuration without any block being in contact with the robot. A single policy PPO does not solve the task and on average has a success rate of $24\%$. The approach with the 1-hot modulation signal achieves a success rate of $42\%$. The options framework stacks the blocks in $54\%$ of episodes on average but takes more time to train. MLSH learns faster than the two previously discussed methods. However, it plateaus out and reaches the same success rate as the options. MPH outperforms all the baselines, both in terms of final average score and the speed of learning. MPH achieves a success rate of $70\%$ on average (5 seeds) and the best random seed stacks the blocks in $98\%$ of episodes. In contrast to the options framework, MPH uses the whole batch to train all the networks and in contrast to MLSH, trains jointly in a single phase always updating all the networks.

\paragraph{Modulation.}
\begin{figure}[t]
\centering
\includegraphics[width=\textwidth]{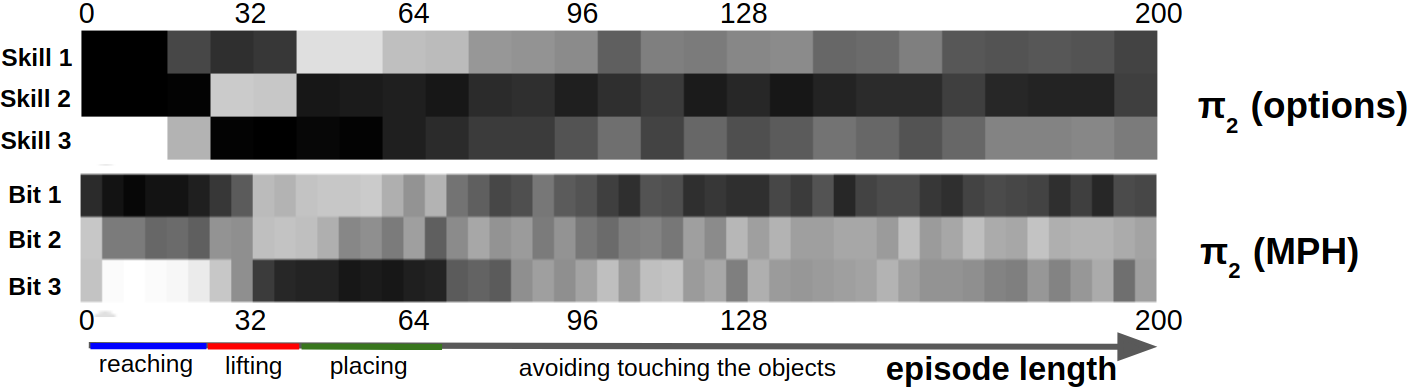}
\caption{Modulation signals for a trained options baseline and MPH on the stacking task, averaged over 50 episodes. The master policy acts every 8 steps for the options framework and 4 steps for the MPH agent, which we obtained by grid search. The options policy can only select one skill at a time, while our MPH uses a bit vector that allows to blend between skills. This results in smoother modulation patterns and higher task performance.}
\label{fig:skills}
\end{figure}

To obtain a better understanding on the role of the modulation signal, we plot histograms of the master policies' decisions, for both the options baseline and MPH. \cref{fig:skills} shows the histogram for a single random seed robustly solving the task. First, we notice that the options master (acting on a time-scale of 8) takes consistent decisions and prefers certain skills over others at each timestep (\cref{fig:skills} top). We highlight the fact that the master policy network is memoryless and does not observe the current timestep value. In the beginning, the master chooses the 3rd skill for about 24 timesteps, then it chooses the 2nd skill for roughly 16 timesteps and finally the first one for the rest of the episode. Thus we conclude that the skills correspond to reaching, lifting and placing primitives which is confirmed by observing the policy acting. Once trained, the options framework solves the problem in the first third of an episode and spends the rest of the time avoiding contact with any block (requires no specific skill). We observe a similar pattern for the MPH modulation signal switching the bits in roughly the same time intervals. The master policy of MPH acts on a time-scale of 4 and changes the modulation signal in roughly the same time intervals as the options master. However, MPH typically employs more than a single bit and benefits from higher modulation skill capacity than the categorical methods like options and MLSH.

\subsection{Pushing}
\label{sec:pushing}

\paragraph{Task description.}

The block pushing task is FetchPush-v1 from OpenAI Gym where following \cite{Andrychowicz2017HindsightER}, we discard initial state-goal pairs in which the goal is already satisfied. In FetchPush-v1, the end-effector is controlled in XYZ space and the goal is to push a randomly placed box to the target. The agent receives the reward of 1 when the block is in an epsilon ball of the episode target and 0 anywhere else. Each episode is randomly initialized in terms of the robot and the block configurations and the target. The length of the episodes is set to 50.

\paragraph{Hyperparameters.}

We use the same set of hyperparameters as described for the stacking task with several exceptions. We adapt the batch size (set to 32 rollouts), the number of training epochs (set to 32), the policies learning rate (set to 0.0001), the value functions learning rate (set to 0.0003), and the discount factor (set to 0.98). For MLSH we change the warmup time to 10, the training time to 50, and the master policy timescale to 10.

\paragraph{Performance.}

We compare MPH with four baselines on FetchPush-v1. We use episode success as performance metrics (averaged over 32 episodes). The pushing is considered to be successful when the block is close to the episode target. As shown in \cref{fig:pushing}, MPH outperforms all the other approaches. A single policy PPO plateaus out after achieving a success rate of $37\%$. The 1-hot hierarchy and options framework on average solve the task with a success rate of $41\%$ and $43\%$ correspondingly. MLSH performs better than the first three methods and achieves a success rate of $45\%$. The options and MLSH take more time to train due to the fact that each option is trained on a sub set of the batch of data. MPH is on average $26\%$ more successful than the best of the baselines and achieves a success rate of $71\%$ on average and the best random seed is able to successfully push the block in $99\%$ of episodes.

\subsection{Hierarchical intrinsic motivation}
\label{sec:motivation}

\begin{figure}[t]
\centering
\begin{subfigure}[t]{.5\textwidth}
\includegraphics[width=\textwidth]{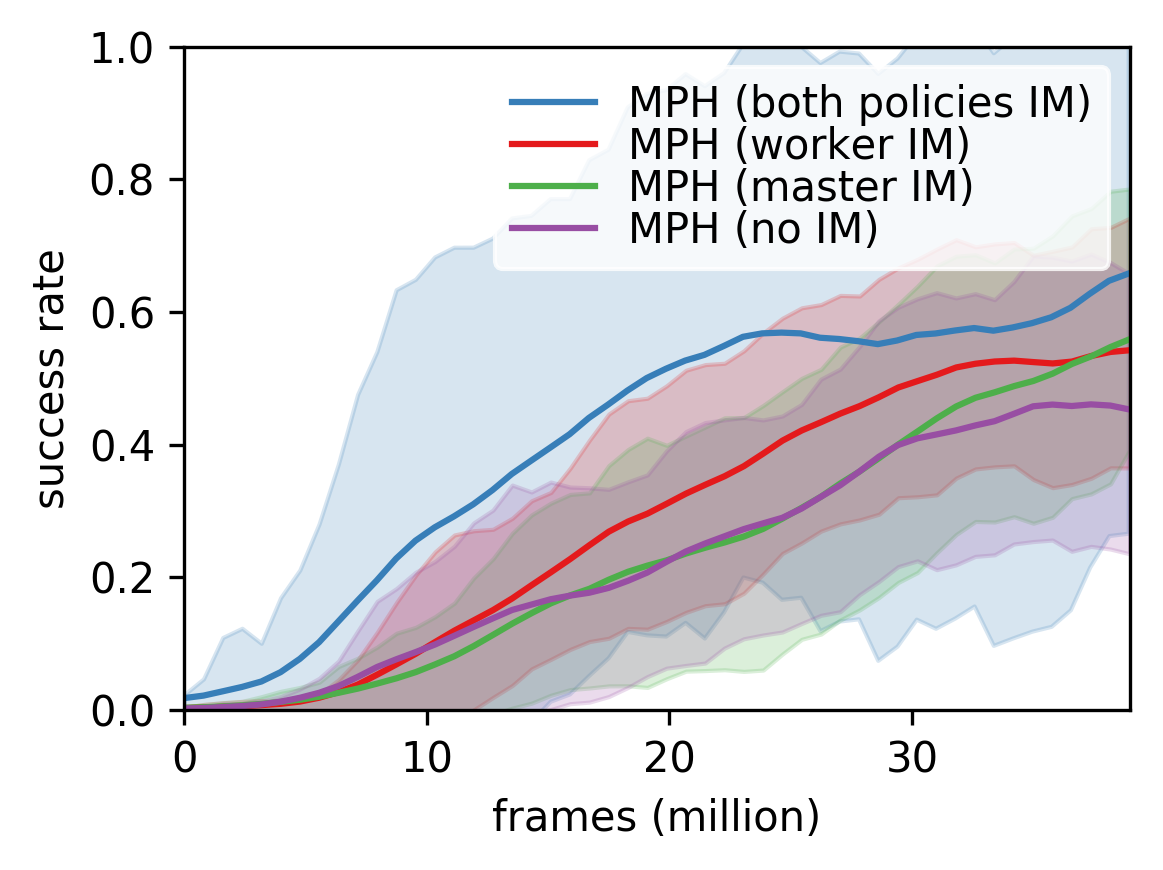}
\caption{Stacking}
\label{fig:stacking_im}
\end{subfigure}\hfill
\begin{subfigure}[t]{.5\textwidth}
\includegraphics[width=\textwidth]{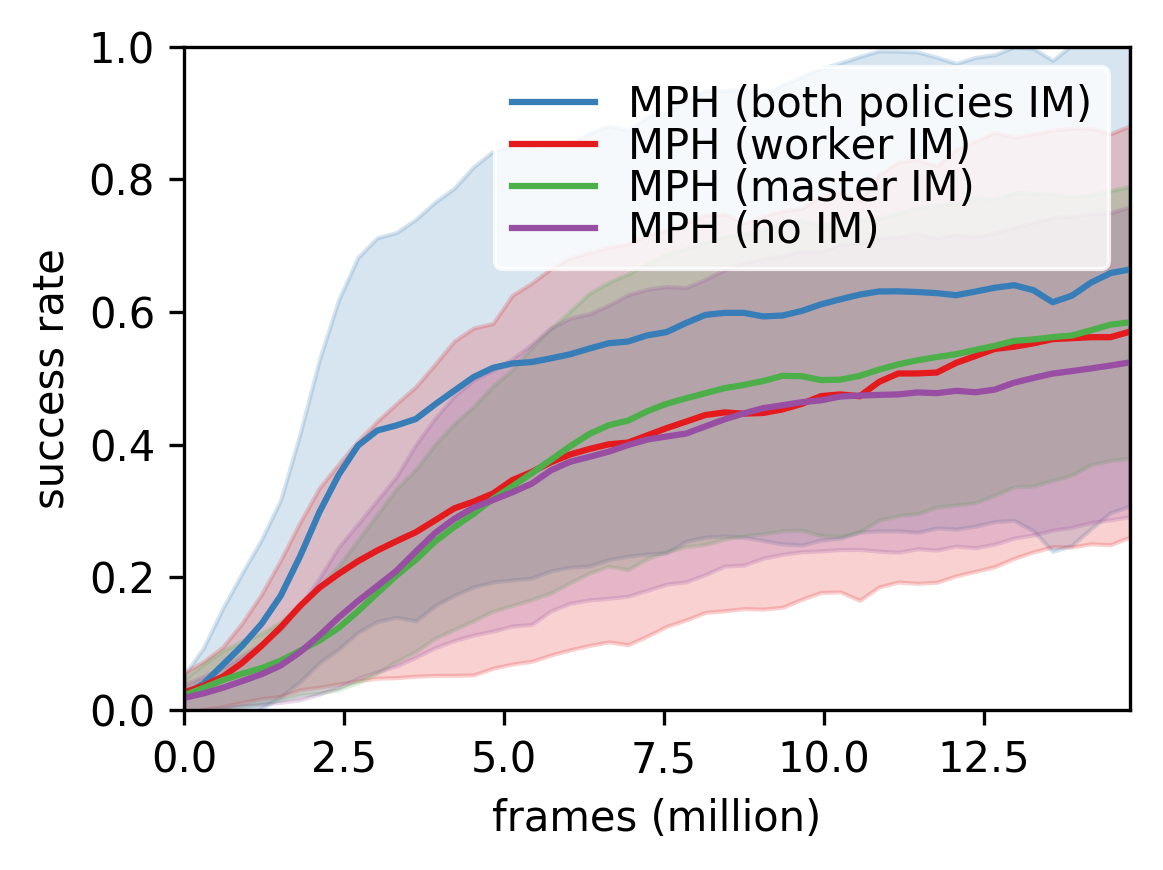}
\caption{FetchPush-v1}
\label{fig:pushing_im}
\end{subfigure}
\caption{Evaluation curves for MPH with and without intrinsic motivation (IM) for each policy. The solid lines correspond to mean success rate, and the shaded lines show the standard deviation. For stacking, both values are mean values for 50 episodes and averaged over top 5 out of 16 random seeds. For pushing, we average over 32 episodes and use 5 random seeds.}
\label{fig:im_ablation}
\end{figure}

We evaluate the effect of intrinsic motivation (IM) applied on both levels of the hierarchy. \Cref{fig:motivation} shows the results for both tasks with the four possible settings of intrinsic rewards: intrinsic reward for both policies, intrinsic reward only for the worker policy, intrinsic reward only for the master policy and no intrinsic reward. We notice that MPH without the intrinsic reward often struggles to find the solution and performs worse. Given the intrinsic bonus for one of the layers, MPH performance improves. The intrinsic motivation on the worker side results in faster initial exploration, however MPH with intrinsically rewarded master network has higher final score, potentially due to better long term planning. The best score is achieved with an intrinsic motivation for both policies. Applying intrinsic motivation to both policies results in an improvement of $20\%$ and $14\%$ w.r.t. the version without intrinsic motivation for the stacking and the pushing tasks correspondingly.

\section{Conclusion}

We introduced Modulated Policy Hierarchies (MPHs) to address environments with sparse rewards that can be decomposed into subtasks. By combing rich modulation signals, temporal abstraction, and intrinsic motivation, MPH benefits from better exploration and increased stability of training. Moreover, in contrast to many state-of-the-art approaches, MPH does not require pre-training, multiple training phases or manual reward shaping. We evaluated MPH on two simulated robot manipulation tasks: pushing and block stacking. In both cases, MPH outperformed baselines and the recently proposed MLSH algorithm, suggesting that our approach may be a fertile direction for further investigation.

\paragraph{Acknowledgements.} This work was supported in part by ERC advanced grant Allegro.

\clearpage
\bibliography{refs}

\clearpage
\appendix
\section{Markovian formulation of MPH}
\label{appendix:derivations}

Since different policies act on different time-scales, we use the following notation for the action and the state of the policy $\pi_k$ acting on the time-scale $T_k$ in an $n$-level hierarchy:
\begin{equation}
    (s_{t+1}^k, a_{t+1}^k) = \begin{cases}
      (s_t^k, a_t^k), & \text{if}\ t \mod T_k \neq 0 \\
      (\{s^\textit{env}_t, a^{k+1}_t, \ldots, a^n_t\}, \pi_k(\{s^\textit{env}_t, a^{k+1}_t, \ldots, a^n_t\})), & \text{otherwise}
    \end{cases}
\end{equation}
We refer to control signals as latent variables of policies on higher levels where the policy on level $k$ is a conditional action distribution $\pi_k(a^k_t | s^k_t)$ assuming that the latent variables $a^{k+1}_t, \ldots, a^n_t$ are the part of the state $s^k_t$. Therefore, the action $a^k_t$ can be sampled once $a^{k+1:n}$ are sampled. The corresponding problem of finding the optimal policy for each hierarchy layer can be solved using the RL machinery once it is reformulated in the MDP formalism as we do below.

For each layer of the hierarchy $k > 1$, we can rewrite the transition probabilities of the MDP on the layer $k$ marginalizing over the actions of the $(k-1)^{th}$ policy and using the transition probabilities of the $(k-1)^{th}$ layer of the hierarchy MDP:
\begin{equation}
    p_k(s^k_{t+1} | s^k_t, a^k_t) = \int_{\mathcal{A}_{k-1}} p_{k-1}(s^{k-1}_{t+1} | s^{k-1}_t, a^{k-1}_t) \pi_{k-1}(a^{k-1}_t | s^{k-1}_t) d a^{k-1}_t
\label{eq:margin_prob}
\end{equation}
Since the higher level policies act on longer time-scales, we also derive the time-scaled transition probabilities for each MDP with the time-scale $T_k > 1$:
\begin{equation}
    p_k(s^k_{t+T_k} | s^k_t, a^k_t) = \overbrace{\int_{\mathcal{A}_k} \ldots \int_{\mathcal{S}_k} \ldots}^{2T_k} p_k(s^k_{t+1}|s^k_t,a^k_t) \prod_{i=1}^{T_k-1} p_k(s^k_{t+i+1} | s^k_{t+i}, a^k_{t+i}) \pi_k(a^k_{t+i} | s^k_{t+i}) d a^k_{t+i} d s^k_{t+i}
\label{eq:timescaled_prob}
\end{equation}
Given \cref{eq:margin_prob} and \cref{eq:timescaled_prob}, one can get the transitions probabilities for any layer MDP using only the environment transition probabilities and the policies. While the former is stationary, the policies are updated in each training epoch that might bring instabilities to the training. The trust region optimization methods, such as TRPO (or its approximation, PPO) bring in a convenient way to bound the changes of the high level MDPs. Such bound can guarantee that the optimization problem solved by TRPO (or PPO) for higher layers changes smoothly during the updates. Thus, the global solution is converging to the optimal solution of the original problem. Below we derive the upper bounds for the transition probabilities change for a discrete case which can be extended to the continuous state and action spaces. We rewrite \cref{eq:margin_prob} for the discrete case ($k$ is shifted by 1):
\begin{equation}
    p_{k+1}(s^{k+1}_{t+1}|s^{k+1}_t, a^{k+1}_t) = \sum_{a^k_t} p_k(s^k_{t+1} | s^k_t, a^k_t) \pi_k(a^k_t | s^k_t) = p_k(s^k_{t+1} | s^k_t, \cdot)^T \pi_k(\cdot|s^k_t)
\end{equation}
We start by deriving the equation for the first two levels of the hierarchy. We denote the transition probability after the training epoch as $p'_k$ and the updated policy as $\pi'_k$. Since $p_1 = p_{\rm{env}} = p'_{\rm{env}} = p'_1$ where $p_{\rm{env}}$ is the transition probability of the environment, we get the following inequalities:
\begin{align}
\label{eq:prob_bound11}
    \abs{p_2(s^2_{t+1}|s^2_t, a^2_t) - p'_2(s^2_{t+1}|s^2_t, a^2_t)} &\leq \norm{p_1(s^1_{t+1} | s^1_t, \cdot)}_{\infty} \norm{\pi_1(\cdot|s^1_t) - \pi'_1(\cdot|s^1_t)}_1 \\
\label{eq:prob_bound12}
    & < \sqrt{\frac{1}{2}D_{KL}(\pi_1(\cdot|s^1_t), \pi'_1(\cdot|s^1_t))} \\
    & \leq \sqrt{\frac{1}{2}\delta_1}
\label{eq:prob_bound13}
\end{align}

where we used Hölder's inequality in \cref{eq:prob_bound11}, Pinsker's inequality and the fact that $p_1 = p_{\rm{env}}$ is ergodic in \cref{eq:prob_bound12} and TRPO (or PPO) guarantee of $D_{KL}(\pi(\cdot|s^1_t), \pi'(\cdot|s^1_t)) < \delta$ in \cref{eq:prob_bound13}.

Next we derive the bound for $k>2$ level of the hierarchy:
\begin{align}
\label{eq:prob_boundk1}
    &\abs{p_k(s^k|s^k_t, a^k_t) - p'_k(s^k_{t+1}|s^k_t, a^k_t)} \\
    &= \left|p_{k-1}(s^{k-1}_{t+1} | s^{k-1}_t, \cdot)^T \pi_{k-1}(\cdot|s^{k-1}_t) - p'_{k-1}(s^{k-1}_{t+1} | s^{k-1}_t, \cdot)^T \pi_{k-1}(\cdot|s^{k-1}_t)\right. \\
    &\left.+ p'_{k-1}(s^{k-1}_{t+1} | s^{k-1}_t, \cdot)^T \pi_{k-1}(\cdot|s^{k-1}_t) - p'_{k-1}(s^{k-1}_{t+1} | s^{k-1}_t, \cdot)^T \pi'_{k-1}(\cdot|s^{k-1}_t) \right| \\
    &\leq \norm{p_{k-1}(s^{k-1}_{t+1} | s^{k-1}_t, \cdot) - p'_{k-1}(s^{k-1}_{t+1} | s^{k-1}_t, \cdot)}_{\infty} \norm{\pi_{k-1}(\cdot|s^{k-1}_t)}_1 \\
    &+ \norm{p'_{k-1}(s^{k-1}_{t+1} | s^{k-1}_t, \cdot)}_{\infty} \norm{\pi_{k-1}(\cdot|s^{k-1}_t) - \pi'_{k-1}(\cdot|s^{k-1}_t)}_1 \\
    &\leq \norm{p_{k-1}(s^{k-1}_{t+1} | s^{k-1}_t, \cdot) - p'_{k-1}(s^{k-1}_{t+1} | s^{k-1}_t, \cdot)}_{\infty} + \norm{\pi_{k-1}(\cdot|s^{k-1}_t) - \pi'_{k-1}(\cdot|s^{k-1}_t)}_1 \\
    & \leq \sum_{i=1}^{k-1}\sqrt{\frac{1}{2}\delta_i}
\label{eq:prob_boundk3}
\end{align}
where we use Hölder's and Pinsker's inequalities and the result of \cref{eq:prob_bound13}.

We showed that for any layer of MPH its MDP's transition probabilities change is upperbounded with the TRPO (or PPO) update. In addition, this bound scales linearly with $k$. Thus, we have a direct control on how much the MDPs on higer layers change after each policy update. Given that the change is small, the optimization problem solved by TRPO (or PPO) for higher layers will also change smoothly during the updates. Therefore, we can apply the standard RL machinery for the hierarchical time-scaled MDPs with the given transition probabilities independently for each layer. Moreover, such guarantees also mean more stable training of the hierarchy.

\end{document}